\documentclass[11pt,a4paper]{article}
\usepackage[hyperref]{emnlp2020}
\usepackage{times}
\usepackage{latexsym}

\newcommand\blfootnote[1]{%
  \begingroup
  \renewcommand\thefootnote{}\footnote{#1}%
  \addtocounter{footnote}{-1}%
  \endgroup
}

\usepackage{microtype}

\aclfinalcopy 

\title{How Cute is Pikachu? Gathering and Ranking Pokémon Properties from Data with Pokémon Word Embeddings}

\author{Mika Hämäläinen, Khalid Alnajjar and Niko Partanen \\
  Department of Digital Humanities \\
  University of Helsinki \\
  \texttt{first.lastname@helsinki.fi}  \\}

\date{}

\begin{document}
\maketitle
\begin{abstract}
We present different methods for obtaining descriptive properties automatically for the 151 original Pokémon. We train several different word embeddings models on a crawled Pokémon corpus, and use them to rank automatically English adjectives based on how characteristic they are to a given Pokémon. Based on our experiments, it is better to train a model with domain specific data than to use a pretrained model. Word2Vec produces less noise in the results than fastText model. Furthermore, we expand the list of properties for each Pokémon automatically. However, none of the methods is spot on and there is a considerable amount of noise in the different semantic models. Our models have been released on Zenodo.
\end{abstract}

\section{Introduction}

Using\blfootnote{This is an English translation of the original paper published in Finnish: Hämäläinen, M.,  Alnajjar, K. \& Partanen, N. (2021). Nettikorpuksen avulla tuotettuja sanavektorimalleja Pokémonien ominaisuuksien kuvaamiseksi. In Saarikivi, T. \& Saarikivi, J. (eds.) \textit{Turhan tiedon kirja -- Tutkimuksista pois jätettyjä sivuja}. p. 199-214. SKS Kirjat.} knowledge-bases that contain properties typical for nouns has been in the heart of computational creativity research for a long time. Such data has proven itself useful when generating a variety of different types of creative language such as metaphors \cite{veale2007comprehending}, poems \cite{hamalainen2018harnessing} or riddles \cite{ritchie2003jape}.

In this paper, we present a novel approach for constructing such a knowledge-base automatically for the 151 original Pokémon. Our approach is applicable in scenarios with a limited amount of data available. The resulting knowledge-base can be used in the future for generating creative language based on Pokémon such as similes and metaphors (e.g. cute as a Pikachu or confused as a Psyduck). We have made the Pokémon word embeddings models\footnote{Pokémon word embeddings models: https://zenodo.org/record/4554478} freely available on Zenodo together with the Pokémon story corpus\footnote{Pokémon corpus: https://zenodo.org/record/4552785}.

Pokémon has been a topic of research in the past \cite{10.1145/3337722.3337739,geisslerpokerator,vaterlaus2019reliving}. However, it has eluded any wide-spread NLP research interest. However, Pokémon names are surprisingly problematic for current NLP methods as we will show in this paper.

Stereotypical knowledge has been successfully extracted in the past \cite{veale2008enriching}. Their method relied on using Google search API to mine stereotypical adjective-noun relations with an "AS \textit{adjective} AS [a/an] \textit{NOUN}" query. However, such a method requires a lot of data in order for it to work and using such a query on a reasonably sized corpus yields hardly any results, based on our experiences.

For proper nouns, or more precisely famous characters, the simplest approach for building such a knowledge-base has been manual annotation as in the case of the Non-Official Characterization list \cite{veale2016round}. While, the NOC list is a valuable resource for properties for famous characters, we are looking into a more automated method for producing a similar knowledge-base for Pokémon.

There has been an automated effort for expanding the properties recorded in the NOC list \cite{al2017expanding}. While this method is a step towards the desired direction in the sense that it does not require the nouns to exist in a massive corpus, it still relies on mined associations between adjectival properties and a hand annotated list of properties for famous characters in order to expand them further.

In our approach, we propose a method for extracting properties for Pokémon automatically from a very small corpus. Furthermore, we use a larger Pokémon specific corpus to automatically rank the extracted properties so that a higher rank is given to the properties that are most descriptive of a given Pokémon.


\begin{table*}[]
\centering
\tabcolsep=0.04cm
\scriptsize
\begin{tabular}{|l|l|l|l|l|l|}
\hline
Pokémon & TF-IDF & Pokémon fastText & Pre-trained fastText & Pokémon Word2Vec & Pokémon Relatedness \\ \hline
Parasect & \begin{tabular}[c]{@{}l@{}}back, big, dark, \\ lower, parasite\end{tabular} & \textbf{\begin{tabular}[c]{@{}l@{}}parasitic, poisonful, \\ sapping, crab-like, poison\end{tabular}} & QF, Oz, EP, XL, foe & \begin{tabular}[c]{@{}l@{}}sapping, crab-like, \\ Polish, poison, sapped\end{tabular} & \begin{tabular}[c]{@{}l@{}}scuttled, solar, evolved,\\ sent, male\end{tabular} \\ \hline
Omanyte & \begin{tabular}[c]{@{}l@{}}full, twisted, pokemon,\\ original, strange\end{tabular} & \begin{tabular}[c]{@{}l@{}}fossil, Oman, Omani, \\ fossil-like, crab-like\end{tabular} & JV, EP, tapu, mi, zoid & \textbf{\begin{tabular}[c]{@{}l@{}}beached, fossil, crab-like, \\ dorsal, evolved\end{tabular}} & \begin{tabular}[c]{@{}l@{}}fossil, caught, scald,\\ level, prehistoric\end{tabular} \\ \hline
Horsea & pokemon, original, powerful & \textbf{\begin{tabular}[c]{@{}l@{}}bubble, squirtish, high-current, \\ splashing, swime\end{tabular}} & QF, ray, zoid, animé, peaty & \begin{tabular}[c]{@{}l@{}}bubble, beached, \\ high-pressured, dorsal, scald\end{tabular} & \begin{tabular}[c]{@{}l@{}}bubble, caught, evolved, \\ level, swimming\end{tabular} \\ \hline
Arcanine & \begin{tabular}[c]{@{}l@{}}pure, true, mysterious, \\ select, majestic\end{tabular} & \begin{tabular}[c]{@{}l@{}}lubric, whinny, mane, \\ canine, dismounted\end{tabular} & EP, XL, JV, pi, glew & \textbf{\begin{tabular}[c]{@{}l@{}}whinny, earth-shaking, \\ orange-yellow, high-pressured, scald\end{tabular}} & \begin{tabular}[c]{@{}l@{}}back, canine, large,\\ sent, male\end{tabular} \\ \hline
Abra & original, psychic & \begin{tabular}[c]{@{}l@{}}disable, Mole, Chinglish, \\ psychic, Minimite\end{tabular} & Oz, ex, D., EP, Ona & \textbf{\begin{tabular}[c]{@{}l@{}}hypnotic, dinged, sapping, \\ psychic, evolved\end{tabular}} & \begin{tabular}[c]{@{}l@{}}psychic, teleporting,\\ side, evolved, cast\end{tabular} \\ \hline
Seaking & \begin{tabular}[c]{@{}l@{}}prominent, pokemon, \\ original\end{tabular} & \begin{tabular}[c]{@{}l@{}}beached, high-current, swime, \\ dorsal, hydro\end{tabular} & QF, JV, EP, A1, zoid & \textbf{\begin{tabular}[c]{@{}l@{}}beached, tidal, high-pressured, \\ seismic, dorsal\end{tabular}} & \begin{tabular}[c]{@{}l@{}}released, trapped, \\ swimming, sent, causing\end{tabular} \\ \hline
Jolteon & \begin{tabular}[c]{@{}l@{}}smallest, negative, sad, \\ shortest, startled\end{tabular} & \begin{tabular}[c]{@{}l@{}}mane, bristled, crackled,\\ wagging, veed\end{tabular} & QF, EP, XL, JV, pi & \begin{tabular}[c]{@{}l@{}}whinny, supercharged, \\ high-pressured, pi, wagging\end{tabular} & \textbf{\begin{tabular}[c]{@{}l@{}}evolved, electric, \\ spiky, male, female\end{tabular}} \\ \hline
Magmar & \begin{tabular}[c]{@{}l@{}}fiery, pokemon, original,\\ smaller, intense\end{tabular} & \begin{tabular}[c]{@{}l@{}}fire-hot, knock-on, punch, \\ seismic, scald\end{tabular} & foe, Oz, EP, XL, zoid & \begin{tabular}[c]{@{}l@{}}five-pointed, high-pressured, \\ seismic, scald, hydro\end{tabular} & \textbf{\begin{tabular}[c]{@{}l@{}}punch, fiery, sent, \\ flame, causing\end{tabular}} \\ \hline
Pidgeot & \begin{tabular}[c]{@{}l@{}}beautiful, top, wide,\\ thick, unsuspecting\end{tabular} & \begin{tabular}[c]{@{}l@{}}bat-wing, cawing, flappish,\\ preened, flapped\end{tabular} & QF, EP, XL, zoid, glew & \textbf{\begin{tabular}[c]{@{}l@{}}cawing, flapped, seismic, \\ lightning-quick, roosting\end{tabular}} & \begin{tabular}[c]{@{}l@{}}back, flapped, evolved, \\ flapping, landed\end{tabular} \\ \hline
\end{tabular}
\caption{Top 5 adjectives produced by different methods for 9 randomly selected Pokémon.}
\label{tab:poke_results}
\end{table*}

\section{Data and Preprocessing}

In order to gather properties for each Pokémon, we look into Wikidata\footnote{https://www.wikidata.org/}, while Wikidata does not contain descriptive properties, it provides us with unambiguous links to Giantbomb\footnote{https://www.giantbomb.com/} entries. We use the Wikidata entry \textit{list of Pokémon introduced in Generation I}\footnote{https://www.wikidata.org/wiki/Q3245450} to obtain these Giantbomb links.

Giantbomb is a website listing information on video game characters. Unlike resources such a Bulbapedia\footnote{https://bulbapedia.bulbagarden.net/} they provide a concise description including useful information such as characteristics and physical abilities without going too deep into the use of the Pokémon in video games. This data, however, is not structural but rather free formed textual description. This data constitutes our small Pokémon description corpus.

In order to rank Pokémon properties, we crawl a larger corpus of texts written about Pokémon. Many Wikipedia-like sources are too neutral to reveal anything meaningful about Pokémon, Pokédex entries are usually too short and non-descriptive for our needs. Subtitles form the Pokémon TV show come with their own problem of audio-visual grounding of the text. Fortunately, we found a great resource of stories authored by Pokémon fans called Fanfiction\footnote{https://www.fanfiction.net/}.

The evident problem of the resource is that many of the stories are poorly written, and that there are stories written in multiple languages. To mitigate this, we use the search functionality of the service to find stories by the query \textit{pokemon} that are in English and have at least 10k words. This results in 8,011 fan-authored stories about Pokémon. We crawl only the stories that meet these criteria. This forms our bigger Pokémon stories corpus, which we process by doing sentence and word tokenization with NLTK \cite{nltk}.

\section{Extracting Pokémon Properties}
We experiment with multiple ways of extracting the properties for each Pokémon. In the first method, we use TF-IDF (term frequency–inverse document frequency) based method for extracting and ranking Pokémon properties on the description corpus. We compare the results of the TF-IDF method to different methods using semantic relatedness and similarity word embeddings models. For semantic relatedness we build a log-likelihood matrix of term-to-term relations based on their co-occurrences following the implementation of Meta4Meaning \cite{xiao2016meta4meaning} and for semantic similarities we utilize word2vec \cite{mikolov2013efficient} and fastText~\cite{bojanowski2016enriching} models. We test out all the methods with generic pretrained models and with domain-specific models trained on the story corpus to see how big of a difference a domain specific corpus makes for the task of automatic extraction of properties.

\begin{table*}[!th]
\centering
\tabcolsep=0.04cm
\scriptsize
\begin{tabular}{|l|l|l|l|l|l|}
\hline
Pokémon & TF-IDF & Pokémon fastText & Pre-trained fastText & Pokémon word2Vec & Pokémon Relatedness \\ \hline
Drowzee & \textbf{\begin{tabular}[c]{@{}l@{}}dangerous, knowledgeable, \\ intelligent, ruthless, twisted\end{tabular}} & \begin{tabular}[c]{@{}l@{}}beautiful, dreamy, \\ raw, alluring, sensuous\end{tabular} &  & \begin{tabular}[c]{@{}l@{}}amusing, funny, surprising, \\ charming, relaxed\end{tabular} & \begin{tabular}[c]{@{}l@{}}public, versatile, despicable,\\ specified, needed\end{tabular} \\ \hline
Magnemite & \begin{tabular}[c]{@{}l@{}}light, inconspicuous, fresh, \\ insignificant, memorable\end{tabular} & \begin{tabular}[c]{@{}l@{}}handsome, rugged, \\ individual\end{tabular} &  & \textbf{\begin{tabular}[c]{@{}l@{}}inflexible, fixed, boring, \\ stolid, unchanging\end{tabular}} & \begin{tabular}[c]{@{}l@{}}soulful, grandiose, expressive, \\ exciting, urgent\end{tabular} \\ \hline
Raichu & \begin{tabular}[c]{@{}l@{}}dismayed, amazed, \\ horrified, outraged, surprised\end{tabular} & \begin{tabular}[c]{@{}l@{}}grandiose, funky, twisty, \\ crazed, exciting\end{tabular} &  & \begin{tabular}[c]{@{}l@{}}grandiose, funky, twisty, \\ crazed, exciting\end{tabular} & \begin{tabular}[c]{@{}l@{}}sturdy, potent, raw, \\ versatile, wealthy\end{tabular} \\ \hline
Beedrill & \begin{tabular}[c]{@{}l@{}}loud, dangerous, \\ clear, deadly, slick\end{tabular} &  &  & \begin{tabular}[c]{@{}l@{}}bitter, divisive, alive,\\ deadly, vulgar\end{tabular} & \begin{tabular}[c]{@{}l@{}}frustrated, disappointed, \\ bitter, scared, shocked\end{tabular} \\ \hline
Exeggcute & \begin{tabular}[c]{@{}l@{}}beautiful, creative, innovative, \\ varied, diverse\end{tabular} & \begin{tabular}[c]{@{}l@{}}satisfying, healthy, safe, \\ delicious, tasty\end{tabular} &  & \begin{tabular}[c]{@{}l@{}}scary, inhuman, cunning, \\ brutal, mean\end{tabular} &  \\ \hline
Weezing & \begin{tabular}[c]{@{}l@{}}creative, innovative, fresh, \\ memorable, quirky\end{tabular} & \textbf{\begin{tabular}[c]{@{}l@{}}harmful, dangerous, deadly, \\ slick, lethal\end{tabular}} & \textbf{} & \textbf{\begin{tabular}[c]{@{}l@{}}harmful, dangerous, deadly, \\ slick, lethal\end{tabular}} & \begin{tabular}[c]{@{}l@{}}dangerous, public, specified, \\ needed, slick\end{tabular} \\ \hline
Meowth & \begin{tabular}[c]{@{}l@{}}beautiful, professional, intelligent,\\ expressive, versatile\end{tabular} & \textbf{\begin{tabular}[c]{@{}l@{}}crafty, clever, funny, \\ well-meaning, treacherous\end{tabular}} &  & cunning, brutal & \begin{tabular}[c]{@{}l@{}}dominant, raised, identifying, \\ normal, known\end{tabular} \\ \hline
Ninetales & \textbf{\begin{tabular}[c]{@{}l@{}}beautiful, shiny, \\ round, passionate, merry\end{tabular}} & \begin{tabular}[c]{@{}l@{}}crafty, clever, funny, \\ well-meaning, treacherous\end{tabular} &  & \begin{tabular}[c]{@{}l@{}}dominant, raised, identifying, \\ normal, known\end{tabular} & \begin{tabular}[c]{@{}l@{}}evocative, natural, fallible, \\ alive, feminine\end{tabular} \\ \hline
Arbok & \begin{tabular}[c]{@{}l@{}}scary, dangerous, funny, \\ intense, ruthless\end{tabular} & \begin{tabular}[c]{@{}l@{}}fluid, dangerous, detestable, \\ unpredictable, totalitarian\end{tabular} &  & \textbf{\begin{tabular}[c]{@{}l@{}}dangerous, potent, odious, \\ slick, carcinogenic\end{tabular}} & \begin{tabular}[c]{@{}l@{}}dominant, feminine, identifying, \\ damp, busted\end{tabular} \\ \hline
\end{tabular}
\caption{Expanded properties for 9 randomly selected Pokémon.}
\label{tab:pokexpand}
\end{table*}

We collect an initial set of adjectival properties for each Pokémon from the Pokémon description corpus by processing it using spaCy\cite{honnibal-johnson:2015:EMNLP} and retaining adjectives appearing in the descriptions of the Pokémon. This step yields an unranked list of few adjectival properties that are used to describe the Pokémon. However, it also includes very generic adjectives such as \textit{original} and in some cases might find no adjectives due to very short descriptions. As an example, the properties collected for \textit{Pikachu} included: \{\textit{electric, petite, close, cute, yellow, high, \ldots, first, electrical}\}.

Next, we investigate methods for ranking and expanding the properties of each Pokémon. The first method makes use of the TF-IDF method where we build the TF-IDF matrix from the Pokémon description corpus by treating Pokémon as documents and their descriptions as features using Scikit-learn~\cite{scikit-learn}. The intuition here is that TF-IDF would capture the importance of each feature to Pokémon. As a result, this gives us a list of words for each Pokémon together with its strength of importance to the Pokémon. This is a very simplistic way of ranking the Pokémon properties without using the larger story corpus. Using the importance scores returned by TF-IDF to rank the properties retrieved in the previous step, we get the following ranked properties to \textit{Pikachu}: \{\textit{lovable, onomatopoetic, prolific, stubborn, superlative, unbeknownst, \ldots, -, 15th}\}.

In the following steps, we rank the collected adjectival properties using semantic relatedness and similarities word embeddings models. For each method, we test out two versions, one that is pretrained on generic text such as Common Crawls and Wikipedia, and another that is trained on the Pokémon stories corpus.

We follow the approach described by~\cite{xiao2016meta4meaning} to build a relatedness matrix by obtaining co-occurrences and then compute the simple log-likelihood as a measurement of relatedness between two words based on their individual frequencies and their observed and expected co-occurrences in the corpus. We use the ukWac corpus~\cite{ferraresi2008introducing} as the generic corpus and build two relatedness models using the generic corpus and the Pokémon stories corpus. It appears that none of the Pokémon got captured in the generic model except for two Pokémon, \textit{Persian} and \textit{Ditto}, which is due to the different meaning they represent in the real world. Ranking \textit{Pikachu} properties using the domain-specific model results in: \{\textit{electric, yellow, electrical, female, quick, powerful, \ldots, exclusive, maximum}\}.

We use word2vec and fastText as the semantic similarity word embeddings models. We use a skip-gram model with the default hyperparameters for both fastText and word2vec. Our word embeddings method consists of having a list of properties (adjectives) the similarity of which is compared against the vector of each Pokémon by a dot product. The more similar the property is to a Pokémon, the higher it ranks.
As the pretrained word2vec and fastText models, we use the models provided by~\cite{kutuzov2017word}\footnote{http://vectors.nlpl.eu/repository/20/3.zip} and~\cite{mikolov2018advances}, respectively. For our Pokémon-specific model, we utilize Gensim \cite{rehurek_lrec} to train the word2vec model and the official fastText library~\cite{bojanowski2017enriching} to build the fastText model from the Pokémon stories corpus.

Similarly to the generic relatedness model, Pokémon names did not appear in the pretrained word2vec model. Nonetheless, due to the fastTexts ability to use subword information during the training phase, it was able to produce semantic similarities between Pokémon and adjectival properties. Sorting \textit{Pikachu}'s properties using the pretrained fastText and Pokémon-specific word2vec and fastText models gives:\\
\indent fastText (pretrained): \{\textit{cute, chuchu, red, -, evil, yellow, japanese, \ldots, tumultuous, non}\},\\
\indent word2vec (Pokémon): \{\textit{electric, chuchu, -, electrical, quick, yellow, cute, \ldots, capable, prominent}\},\\
\indent fastText (Pokémon): \{\textit{electric, chuchu, electrical, cute, yellow, close, \ldots, prolific, 15th}\}.

In order to extract a ranked list of properties for each Pokémon from the word embedding models, we compute the similarity for each Pokémon and every single adjective in the Oxford English Dictionary (OED)\footnote{https://www.oed.com/} and sort these words (properties) based on their similarity with each Pokémon.


Furthermore, we experiment with an existing method for expanding properties for the results of each method. The property expansion is based on the data and algorithm presented by \citeauthor{al2017expanding} (\citeyear{al2017expanding}). The method takes in a list of properties and produces an extended property list by using Thesaurus Rex data \cite{veale2013creating}. We use this method to predict more properties by feeding in the top 10 adjectives produced by each model.

\section{Results}

Table \ref{tab:poke_results} shows results for different Pokémon by the different methods. The table shows results for the word embeddings models when using adjectives from the OED. The pretrained Word2Vec model and generic relatedness model are missing from the table as they did not produce any results at all for any Pokémon. The cells in bold have the highest number of descriptive adjectives.

We can see that the pre-trained fastText model does not capture the semantics of any Pokémon at all. All in all, fastText seems to produce good adjectives in the top results, but it clearly struggles with the out-of-vocabulary adjectives. Instead of not returning a vector for them at all, and thus ignoring them, it has been designed to return vectors based on the character level similarity of the word. For this reason, \textit{Oman} and \textit{Omani}, words that did not occur in the training corpus, get highly associated with Omanyte, as their character distance is low. For water Pokémon, \textit{swime}\footnote{OED: Used vaguely (like the noun) in Destr. Troy = giddy, dazed, and (actively) stunning.} gets a high score mostly due to the fact that it is close to the word \textit{swim}.

Throughout the results, we can see that the obscurity of some of the adjectives in the OED confuses the models. Better results could be achieved if the list of adjectives was obtained from a corpus instead of a comprehensive dictionary that also records historical, obsolete and dialectal words.

It is very difficult to pick the overall best model for the task, as all of them work better for certain Pokémon than the others. We can, however, gather that word embedding models that are trained on a domain specific corpus work better than using TF-IDF to extract terms from short documents or using a pretrained model. Word2Vec seems to produce less noise than fastText.

In Table \ref{tab:pokexpand}, we can see the resulting top 5 new properties produced by the automatic expansion of properties based on the lists for the top 10 properties produced by each method. None of the extended properties for Beedrill, Exeggcute and Raichu were descriptive enough to be highlighted as the best result. All in all, the expanded properties are very poor at describing each individual Pokémon. Based on these results, we cannot recommend using an automatic property expansion for Pokémon as it seems to favor properties typical for people. The method also failed to expand some of the properties for some models, and all of the properties for the pre-trained fastText model.

\section{Conclusions}

In this paper, we have presented our initial approaches in mining properties for Pokémon characters. The result look promising, although they reveal problems in the semantic representations of word embedding models, especially in pre-trained ones that belong to a different domain of text. The task of automatically extracting meaningful properties is far from trivial and calls for more future work. Nonetheless, our approach is a step away from expert annotated data into a fully automatic methodology.

The journey has just begun, so in the future different experiments could be conducted in terms of what kind of adjectives are used to query the word embedding models for each Pokémon. Also, a hybrid approach could be taken to combine the strengths of each individual model; the more models point towards a certain property, the more likely it is to be a descriptive one of a given Pokémon.

Based on our research, we can conclude that the pretrained models do not work with Pokémon at all. Clearly, Pokémon itself is by no means so deviant a phenomenon that it could not be modeled with word embeddings. The problem we can see is part of a wider phenomenon that has received a little attention in the field of NLP. If pretrained models, which are constantly used in various NLP studies, are not able to describe Pokémon, what other phenomena might they describe equally poorly? In general, our discipline does not pay very much attention to how well computational models work when applied to a completely new context.

The embeddings trained in this paper may be useful in a variety of different computational creativity tasks relating to Pokémon. Therefore we have released the models and the code freely on Zenodo (links on the first page of this paper). 

\bibliographystyle{acl_natbib}
\bibliography{iccc}

\begin{thebibliography}{21}
\expandafter\ifx\csname natexlab\endcsname\relax\def\natexlab#1{#1}\fi

\bibitem[{Alnajjar et~al.(2017)Alnajjar, H{\"a}m{\"a}l{\"a}inen, Chen, and
  Toivonen}]{al2017expanding}
Khalid Alnajjar, Mika H{\"a}m{\"a}l{\"a}inen, Hanyang Chen, and Hannu Toivonen.
  2017.
\newblock Expanding and weighting stereotypical properties of human characters
  for linguistic creativity.
\newblock In \emph{ICCC}, pages 25--32.

\bibitem[{Bird et~al.(2009)Bird, Klein, and Loper}]{nltk}
Steven Bird, Ewan Klein, and Edward Loper. 2009.
\newblock \emph{Natural Language Processing with Python}, 1st edition.
\newblock O’Reilly Media, Inc.

\bibitem[{Bojanowski et~al.(2016)Bojanowski, Grave, Joulin, and
  Mikolov}]{bojanowski2016enriching}
Piotr Bojanowski, Edouard Grave, Armand Joulin, and Tomas Mikolov. 2016.
\newblock Enriching word vectors with subword information.
\newblock \emph{arXiv preprint arXiv:1607.04606}.

\bibitem[{Bojanowski et~al.(2017)Bojanowski, Grave, Joulin, and
  Mikolov}]{bojanowski2017enriching}
Piotr Bojanowski, Edouard Grave, Armand Joulin, and Tomas Mikolov. 2017.
\newblock Enriching word vectors with subword information.
\newblock \emph{Transactions of the Association for Computational Linguistics},
  5:135--146.

\bibitem[{Ferraresi et~al.(2008)Ferraresi, Zanchetta, Baroni, and
  Bernardini}]{ferraresi2008introducing}
Adriano Ferraresi, Eros Zanchetta, Marco Baroni, and Silvia Bernardini. 2008.
\newblock Introducing and evaluating ukwac, a very large web-derived corpus of
  english.
\newblock In \emph{Proceedings of the 4th Web as Corpus Workshop (WAC-4) Can we
  beat Google}, pages 47--54.

\bibitem[{Geissler et~al.(2020)Geissler, Nguyen, Theodorakopoulos, and
  Gatti}]{geisslerpokerator}
Dominique Geissler, Elisa Nguyen, Daphne Theodorakopoulos, and Lorenzo Gatti.
  2020.
\newblock Pokérator-unveil your inner pokémon.
\newblock In \emph{Proceedings of the Eleventh International Conference on
  Computational Creativity}.

\bibitem[{H{\"a}m{\"a}l{\"a}inen(2018)}]{hamalainen2018harnessing}
Mika H{\"a}m{\"a}l{\"a}inen. 2018.
\newblock Harnessing nlg to create finnish poetry automatically.
\newblock In \emph{International Conference on Computational Creativity}, pages
  9--15. Association for Computational Creativity (ACC).

\bibitem[{Honnibal and Johnson(2015)}]{honnibal-johnson:2015:EMNLP}
Matthew Honnibal and Mark Johnson. 2015.
\newblock \href {https://aclweb.org/anthology/D/D15/D15-1162} {An improved
  non-monotonic transition system for dependency parsing}.
\newblock In \emph{Proceedings of the 2015 Conference on Empirical Methods in
  Natural Language Processing}, pages 1373--1378, Lisbon, Portugal. Association
  for Computational Linguistics.

\bibitem[{Kutuzov et~al.(2017)Kutuzov, Fares, Oepen, and
  Velldal}]{kutuzov2017word}
Andrei Kutuzov, Murhaf Fares, Stephan Oepen, and Erik Velldal. 2017.
\newblock Word vectors, reuse, and replicability: Towards a community
  repository of large-text resources.
\newblock In \emph{Proceedings of the 58th Conference on Simulation and
  Modelling}, pages 271--276. Link{\"o}ping University Electronic Press.

\bibitem[{Mikolov et~al.(2013)Mikolov, Chen, Corrado, and
  Dean}]{mikolov2013efficient}
Tomas Mikolov, Kai Chen, Greg Corrado, and Jeffrey Dean. 2013.
\newblock Efficient estimation of word representations in vector space.
\newblock \emph{arXiv preprint arXiv:1301.3781}.

\bibitem[{Mikolov et~al.(2018)Mikolov, Grave, Bojanowski, Puhrsch, and
  Joulin}]{mikolov2018advances}
Tomas Mikolov, Edouard Grave, Piotr Bojanowski, Christian Puhrsch, and Armand
  Joulin. 2018.
\newblock Advances in pre-training distributed word representations.
\newblock In \emph{Proceedings of the International Conference on Language
  Resources and Evaluation (LREC 2018)}.

\bibitem[{Pedregosa et~al.(2011)Pedregosa, Varoquaux, Gramfort, Michel,
  Thirion, Grisel, Blondel, Prettenhofer, Weiss, Dubourg, Vanderplas, Passos,
  Cournapeau, Brucher, Perrot, and Duchesnay}]{scikit-learn}
F.~Pedregosa, G.~Varoquaux, A.~Gramfort, V.~Michel, B.~Thirion, O.~Grisel,
  M.~Blondel, P.~Prettenhofer, R.~Weiss, V.~Dubourg, J.~Vanderplas, A.~Passos,
  D.~Cournapeau, M.~Brucher, M.~Perrot, and E.~Duchesnay. 2011.
\newblock Scikit-learn: Machine learning in {P}ython.
\newblock \emph{Journal of Machine Learning Research}, 12:2825--2830.

\bibitem[{{\v R}eh{\r u}{\v r}ek and Sojka(2010)}]{rehurek_lrec}
Radim {\v R}eh{\r u}{\v r}ek and Petr Sojka. 2010.
\newblock {Software Framework for Topic Modelling with Large Corpora}.
\newblock In \emph{{Proceedings of the LREC 2010 Workshop on New Challenges for
  NLP Frameworks}}, pages 45--50, Valletta, Malta. ELRA.

\bibitem[{Ritchie(2003)}]{ritchie2003jape}
Graeme Ritchie. 2003.
\newblock The jape riddle generator: technical specification.
\newblock \emph{Institute for Communicating and Collaborative Systems}.

\bibitem[{Salter et~al.(2019)Salter, Stanfill, and
  Sullivan}]{10.1145/3337722.3337739}
Anastasia Salter, Mel Stanfill, and Anne Sullivan. 2019.
\newblock \href {https://doi.org/10.1145/3337722.3337739} {But does pikachu
  love you? reproductive labor in casual and hardcore games}.
\newblock In \emph{Proceedings of the 14th International Conference on the
  Foundations of Digital Games}, New York, NY, USA. Association for Computing
  Machinery.

\bibitem[{Vaterlaus et~al.(2019)Vaterlaus, Frantz, and
  Robecker}]{vaterlaus2019reliving}
J~Mitchell Vaterlaus, Kala Frantz, and Tracey Robecker. 2019.
\newblock “reliving my childhood dream of being a pok{\'e}mon trainer”: An
  exploratory study of college student uses and gratifications related to
  pok{\'e}mon go.
\newblock \emph{International Journal of Human--Computer Interaction},
  35(7):596--604.

\bibitem[{Veale(2016)}]{veale2016round}
Tony Veale. 2016.
\newblock Round up the usual suspects: Knowledge-based metaphor generation.
\newblock In \emph{Proceedings of the Fourth Workshop on Metaphor in NLP},
  pages 34--41.

\bibitem[{Veale and Hao(2007)}]{veale2007comprehending}
Tony Veale and Yanfen Hao. 2007.
\newblock Comprehending and generating apt metaphors: a web-driven, case-based
  approach to figurative language.
\newblock In \emph{AAAI}, volume 2007, pages 1471--1476.

\bibitem[{Veale and Hao(2008)}]{veale2008enriching}
Tony Veale and Yanfen Hao. 2008.
\newblock Enriching wordnet with folk knowledge and stereotypes.
\newblock In \emph{Proceedings of the 4th Global WordNet Conference, Szeged,
  Hungary}.

\bibitem[{Veale and Li(2013)}]{veale2013creating}
Tony Veale and Guofu Li. 2013.
\newblock Creating similarity: Lateral thinking for vertical similarity
  judgments.
\newblock In \emph{Proceedings of the 51st Annual Meeting of the Association
  for Computational Linguistics (Volume 1: Long Papers)}, pages 660--670.

\bibitem[{Xiao et~al.(2016)Xiao, Alnajjar, Granroth-Wilding, Agres, and
  Toivonen}]{xiao2016meta4meaning}
Ping Xiao, Khalid Alnajjar, Mark Granroth-Wilding, Kat Agres, and Hannu
  Toivonen. 2016.
\newblock Meta4meaning: Automatic metaphor interpretation using corpus-derived
  word associations.
\newblock In \emph{Proceedings of the 7th International Conference on
  Computational Creativity (ICCC). Paris, France}.

\end{thebibliography}

\end{document}